\pdfoutput=1
\documentclass[11pt]{article}

\usepackage[final]{acl}

\usepackage{times}
\usepackage{latexsym}

\usepackage[T1]{fontenc}
\usepackage[utf8]{inputenc}

\usepackage{microtype}

\usepackage{inconsolata}

\usepackage{graphicx}

\usepackage{xcolor}
\usepackage{hyperref}
\usepackage{cleveref}
\usepackage{booktabs}
\usepackage{adjustbox}
\usepackage{kotex}
\usepackage{multirow}
\usepackage[normalem]{ulem}
\usepackage{amssymb}

\usepackage{lipsum}
\usepackage{soul}

\useunder{\uline}{\ul}{}

\usepackage{colortbl}
\usepackage{arydshln}
\crefformat{section}{\S#2#1#3}
\crefformat{subsection}{\S#2#1#3}
\crefformat{subsubsection}{\S#2#1#3}

\newcommand\coloredsout[2][red]{{\color{#1}\sout{#2}}}

\newcommand\coloreduwave[2][blue]{\leavevmode\rlap{\uwave{\phantom{#2}}}\textcolor{#1}{#2}}
\title{Evaluating Automatic Speech Recognition Systems\\ for Korean Meteorological Experts}

\author{ChaeHun Park  \hspace{0.5cm} Hojun Cho \hspace{0.5cm} Jaegul Choo \\
  KAIST AI \\
  \texttt{\{ddehun,hojun.cho,jchoo\}@kaist.ac.kr} \\
}

\begin{document}
\maketitle
\begin{abstract}
Automatic speech recognition systems often fail on specialized vocabulary in tasks such as weather forecasting. To address this, we introduce an evaluation dataset of Korean weather queries. The dataset was recorded by diverse native speakers following pronunciation guidelines from domain experts and underwent rigorous verification. Benchmarking both open-source models and a commercial API reveals high error rates on meteorological terms. We also explore a lightweight text-to-speech-based data augmentation strategy, yielding substantial error reduction for domain-specific vocabulary and notable improvement in overall recognition accuracy. Our dataset is available at \url{https://huggingface.co/datasets/ddehun/korean-weather-asr}.
\end{abstract}

\section{Introduction}
Meteorologists rely on vast, complex databases for forecasting, yet crafting precise SQL queries demands specialized expertise. Natural language interfaces address this gap by translating user questions into database queries \citep{zhongSeq2SQL2017,kim2020natural,deng2022recent}. Notably, \citet{jo-etal-2023-integrated} developed an integrated search system for Korean weather data, enabling users to query extensive meteorological information using natural language. 
This system simplifies data retrieval and thereby improves operational efficiency.

Incorporating Automatic Speech Recognition (ASR) into these systems can further enhance usability by enabling voice-driven queries, a practical advantage for busy forecasters. However, off-the-shelf ASR models—often trained on general-domain or English-centric corpora—tend to misrecognize Korean meteorology terms due to both the language’s agglutinative structure and specialized vocabulary \citep{lee2019adversarially,cho-etal-2020-towards,li2021scaling,yadav-sitaram-2022-survey,whisper,ferraz2024distilwhisper,song2024lora}.  

\begin{figure}[t!]
\centering
\includegraphics[width=\columnwidth]{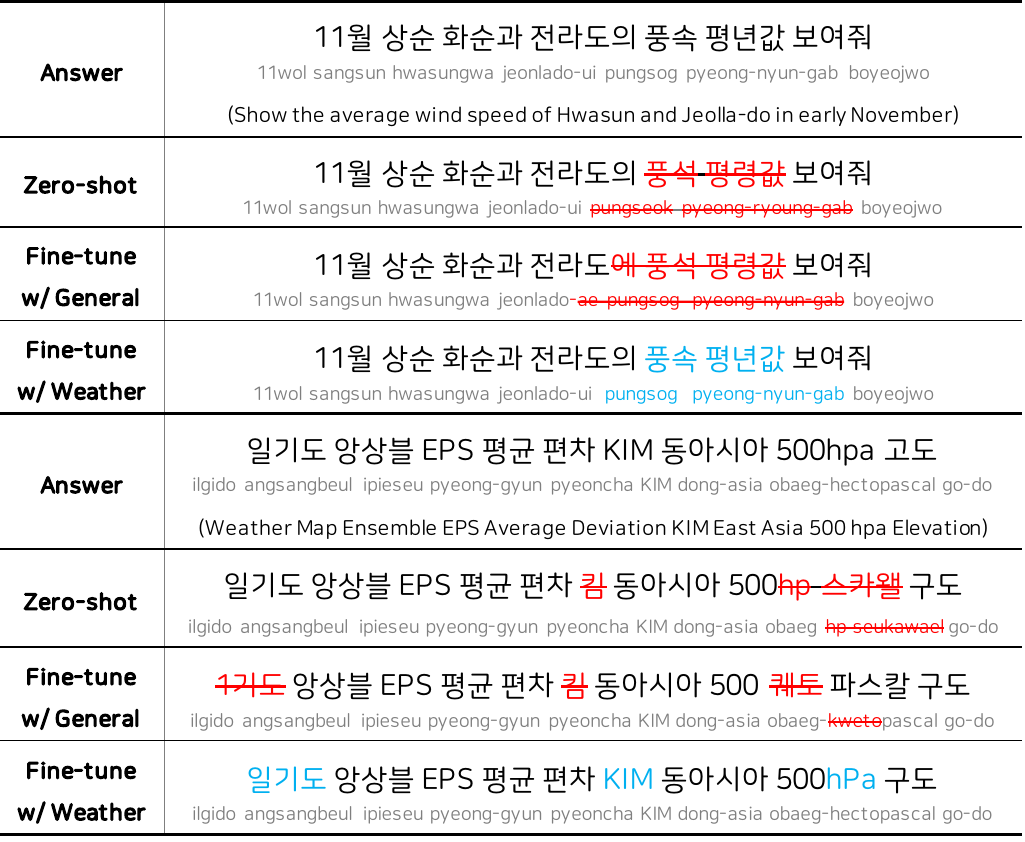}
\caption{\textbf{Qualitative comparison of different models for Korean meteorological queries.}  All models are based on \texttt{Whisper large-v2} \citep{whisper}. The models compared include \texttt{Zero-shot}, fine-tuned on a \texttt{General} domain dataset (i.e., KsponSpeech \citep{bang2020ksponspeech}), and fine-tuned on a \texttt{Weather} domain dataset synthetically generated by a text-to-speech API. Wrongly predicted words are manually highlighted in \textcolor{red}{\st{strikethrough}} by the authors.
}
\label{fig:motif}
\vspace{-0.5cm}
\end{figure}

To quantify these challenges, we created an evaluation set of 5,500 Korean weather queries. Eleven native speakers recorded queries sourced from \citet{jo-etal-2023-integrated}, following expert pronunciation guidelines and a thorough manual validation process to guarantee correctness. We then benchmarked several popular multilingual ASR models (e.g., Whisper variants) and observed consistent misrecognitions of domain-specific terms—such as \textit{weather map}, \textit{average wind speed}, \textit{KIM} (\textit{Korean Integrated Model}), and units like \textit{hPa}—even after fine-tuning on a large general-domain Korean speech corpus \citep{bang2020ksponspeech} (Fig.~\ref{fig:motif}).
% To resolve this issue, we employed a text-to-speech-based data augmentation method, which improved the recognition of domain-specialized terminologies. 
We also explored lightweight approaches on our dataset, including TTS-based data augmentation \citep{zheng2021using} and LLM-based post-processing \citep{hu-etal-2024-listen}.
% We believe our findings provide valuable insights for developing effective domain-specific ASR systems, particularly within the Korean weather forecasting domain.
% Our contributions can be summarized as follows:
% \begin{itemize}
%     \item We developed a specialized evaluation dataset for Korean weather-related ASR tasks by recording and validating spoken queries from native Korean speakers.
 
%     \item We conducted evaluations using different model configurations, identifying areas where performance could be improved with domain-specific dataset creation.
    
%     \item We proposed and demonstrated the effectiveness of a text-to-speech-based data augmentation method to improve ASR performance in recognizing specialized meteorological terms.
% \end{itemize}

Our contributions are as follows: (1) We construct a domain-specific ASR evaluation dataset by recording and validating weather-related queries from native Korean speakers; (2) we evaluate various ASR configurations and highlight the need for domain-specific adaptation; and (3) we explore TTS-based data augmentation to improve recognition of specialized meteorological terms.

\begin{figure}[t!]
\centering
\includegraphics[width=0.9\columnwidth]{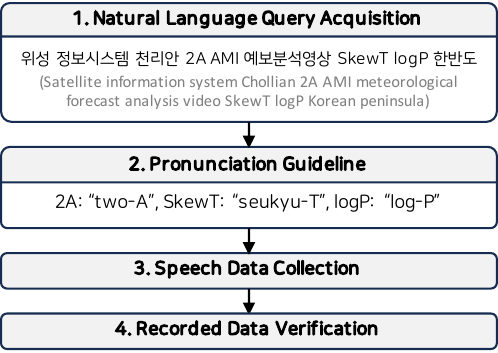}
\caption{A construction pipeline of ASR evaluation dataset for Korean meteorological domain.}
\label{fig:pipeline}
\vspace{-0.3cm}
\end{figure}

\section{Related Work}
\paragraph{Speech Recognition for Specialized Domains}
Speech recognition technology has been extensively studied in various specialized domains, such as medical \citep{le2024vietmed} and financial fields \citep{o2021spgispeech}.
One major challenge is accurately handling domain-specific terminology.
\citet{zheng2021using} demonstrated that synthetic speech data enhances the recognition of out-of-vocabulary words in domain-specific contexts.
\citet{shamsian2024keyword} propose a keyword-guided adaptation to improve the recognition accuracy of models for specialized terms.
Our research specifically focuses on the Korean meteorological domain, creating an evaluation dataset tailored to assess ASR performance in this specialized field.

\paragraph{Speech Recognition for Korean}
Research on ASR for non-English languages, including Korean, presents unique challenges.
Multilingual ASR models, such as Whisper \citep{whisper}, often underperform with languages like Korean due to unique linguistic characteristics and limited representation in training datasets \citep{li2021scaling,yadav-sitaram-2022-survey,ferraz2024distilwhisper,song2024lora}.
This performance drop is often attributed to the lower proportion and diversity of non-English data in training sets.
Efforts to improve ASR for Korean include datasets like \texttt{KsponSpeech}, a spontaneous speech corpus, and \texttt{ClovaCall}, a goal-oriented dialog speech corpus, provide valuable resources for developing and evaluating Korean ASR systems \citep{bang2020ksponspeech,ha2020clovacall}. 
We evaluate ASR performance specifically within the Korean meteorological domain, using a tailored evaluation dataset to identify and address domain-specific performance limitations.

\section{Dataset Construction}
\label{sec:data-overview}
This study aims to develop a specialized dataset for ASR systems within the Korean meteorological domain.
The construction process includes several key steps: Acquiring the natural language questions, obtaining pronunciation guidelines for specialized terms from domain experts, collecting speech data from native speakers, and verifying the correctness of the recorded audio. 
The detailed data construction pipeline (\cref{sec:data-pipeline}, Fig.~\ref{fig:pipeline}) and the analysis of the resulting dataset (\cref{sec:data-analysis}) are as follows.

\begin{table}[t]
\centering
\small
% \begin{adjustbox}{width=0.65\columnwidth}
\begin{tabular}{lcc}
\toprule
Num. samples && 5,500 \\
Utterance time (sec)  && 7.05$_{2.55}$   \\
\,\, - min / max time   &       & 0.92 / 29.98          \\
Avg. chars / words        &&24.49$_{15.62}$ / 7.59$_{4.34}$   \\
% Avg.         &&  \\
Unique words      && 4,955          \\
Absent ratio (\%) &&24.86         \\
\bottomrule
\end{tabular}
% \end{adjustbox}
\vspace{-0.1cm}
\caption{\textbf{Dataset Statistics}. 
% \textit{Uttr. time (sec)} denotes the audio lengths. 
% Words are tokenized using the NLTK library \citep{bird-loper-2004-nltk}. 
\textit{Absent ratio (\%)} refers to the percentage of unique words not presented in the general Korean ASR dataset \citep{bang2020ksponspeech}, divided by the total number of unique words in our dataset. 
% Numbers in subscript represent standard deviation values.
}

\vspace{-0.4cm}
\label{tab:statistics}
\end{table}

\subsection{Data Construction Pipeline}
\label{sec:data-pipeline}
\noindent{\textbf{Acquiring Natural Language Questions}}
We used the natural language question dataset released by \citet{jo-etal-2023-integrated}, which covers diverse meteorological queries. From its URL and SQL subsets, we selected 3,575 and 1,925 questions, respectively, to ensure broad topic coverage. While the original dataset includes structured query mappings (e.g., SQL), we used only the natural language questions, focusing on how forecasters would verbally express their information needs. For more on the original dataset, see \citet{jo-etal-2023-integrated}.

% We began by utilizing the natural language question dataset released by \citet{jo-etal-2023-integrated}.
% This dataset includes a diverse range of questions relevant to the Korean meteorological domain, providing a robust foundation for our spoken query dataset.
% The dataset consists of two subsets: URL and SQL.
% We selected 3,575 questions from the URL subset and 1,925 questions from the SQL subset to ensure comprehensive coverage of various meteorological topics that forecasters typically seek.
% Note that the original dataset in \citet{jo-etal-2023-integrated} contains natural language queries mapped to their corresponding structured queries (e.g., SQL).
% For our purposes, we used only natural language queries, focusing on how these questions would be articulated by users in real-world scenarios. 
% For more details about natural language questions, please refer to the original paper of the dataset \citep{jo-etal-2023-integrated}.

\noindent{\textbf{Obtaining Pronunciation Guidelines from Domain Experts}}
% To ensure the accurate pronunciation of specialized meteorological terminology, we collaborated with domain experts. These experts provided comprehensive guidelines on how to pronounce specific terms correctly. For example, the acronym \texttt{"BUFR"} should be pronounced as \texttt{"buffer"} and\texttt{ "AMI"} as \texttt{"A-M-I."} Additionally, terms such as \texttt{"33009 station"} could be pronounced as \texttt{"three three zero zero nine station"} or \texttt{"thirty-three thousand and nine station."} These detailed instructions were crucial for maintaining the linguistic integrity of the dataset, ensuring that all domain-specific terms were articulated accurately by the speakers. 
% More detailed guidelines are provided in Appendix \ref{app:sec:guideline}.
To ensure the correct pronunciation of domain-specific terms, we collaborated with meteorological experts. They provided detailed guidelines—for example, pronouncing \texttt{"BUFR"} as \texttt{"buffer"} and \texttt{"AMI"} as \texttt{"A-M-I"}. Additionally, terms such as \texttt{"33009 station"} could be pronounced as \texttt{"three three zero zero nine station"} or \texttt{"thirty-three thousand and nine station."} These guidelines helped speakers produce consistent and intelligible renditions of specialized terms. Full details are in Appendix~\ref{app:sec:guideline}.

\begin{table}[t]
\centering
\small
\begin{adjustbox}{width=\columnwidth}
\begin{tabular}{l|c}
\toprule
\multirow{3}{*}{\textbf{General}}&  thing (거) \;\; but (근데) \;\; not (안)  \\
                                 &  that (그) \;\; just (그냥) \;\; I (내가)  \\
                                 &  such (그런) \;\; uh (어)  \;\; what (뭐)   \\

\midrule
                
\multirow{4}{*}{\textbf{Weather}}& show (보여줘) \, AWS (AWS)\\
                                 & inform (알려줘)   \, standard normals (평년값) \\
                                 & surface (지상) \, highest temperature (최고 기온) \,\\
                                 & ocean (해양) \, radar (레이더)  \,weather chart (일기도) \\                                 
\bottomrule
\end{tabular}
\end{adjustbox}
\caption{Nine most frequently occurring words in Korean ASR datasets in open-domain dialog (\texttt{General}) and weather (\texttt{Weather}) domains. The KsponSpeech \citep{bang2020ksponspeech} is used as a general domain dataset. 
% Each English word is manually translated by the authors from the original Korean word presented in the bracket.
}

\label{tab:frequent}
\vspace{-0.4cm}
\end{table}

\noindent{\textbf{Collecting Speech Data from Native Speakers}}
We recruited eleven native Korean speakers (7 male, 4 female, all in their twenties) to record the queries. Participants were provided with the pronunciation guidelines and detailed instructions. Each was compensated at a rate of 16,000 KRW (approx. 11.6 USD) per 100 queries. All recordings were conducted in quiet and controlled environments.

\noindent{\textbf{Verifying the Recorded Voices}}
Following the collection of speech data, a rigorous verification process was implemented.
Human verifiers listened to each recording to ensure accuracy and adherence to the pronunciation guidelines.
They also checked for clarity and the absence of background noise or errors. 
Two of the authors are employed for this verification process.
This process was essential to guarantee the quality and reliability of the dataset for subsequent ASR evaluation.

\begin{table*}[t!]
\centering
\small
\begin{adjustbox}{width=2\columnwidth}
\begin{tabular}{lc|ccc|ccc|ccc}
\toprule
\multirow{2}{*}{\textbf{Model}} & \multirow{2}{*}{\textbf{Params.}} & \multicolumn{3}{c}{\textbf{KsponSpeech}$_{\mathrm{Eval}}$ (3k)}                   & \multicolumn{3}{c}{\textbf{Weather}$_{\mathrm{Dev}}$ (0.5k)}           & \multicolumn{3}{c}{\textbf{Weather}$_{\mathrm{Test}}$ (5.0k)}          \\
\cmidrule(lr){3-5} \cmidrule(lr){6-8} \cmidrule(lr){9-11}
       
&     & CER                 & WER            & sWER           & CER            & WER            & sWER           & CER            & WER            & sWER           \\ 
\hline
\multicolumn{11}{l}{\cellcolor[HTML]{EFEFEF}\textit{ (1) Zero-shot Evaluation of Whisper model family} \citep{whisper}} \\ \hline
Tiny&  39M & 32.67&53.69&47.78&33.53&74.93&54.19&38.24&81.99&58.61\\
Small& 244M &16.75&32.57&26.2&16.39&46.62&27.69&22.92&57.4&34.65 \\
Medium& 769M &14.33&29.73&21.21&12.65&38.67&19.31&19.26&50.38&26.54 \\
Large-v2& 1550M & 14.66&29.97&20.75&11.78&36.31&19.03&18.35&46.74&25.46\\ \hline
\multicolumn{11}{l}{\cellcolor[HTML]{EFEFEF}\textit{(2) Whisper-Large-v2 finetuned on Different Datasets}} \\ \hline
General&  1550M &\textbf{10.48}&\underline{24.05}&\textbf{16.49}&16.92&49.59&21.75&24.57&63.28&28.22 \\
Weather$^*$& 1550M & 15.81&32.48&25.28&\textbf{6.93}&\textbf{17.99}&\textbf{9.13}&\textbf{10.64}&\textbf{31.81}&\textbf{13.61}\\
General+Weather$^*$& 1550M &\underline{10.54}&\textbf{23.94}&\underline{16.52}&\underline{8.68}&\underline{22.58}&\underline{12.50}&\underline{17.46}&\underline{43.41}&\underline{20.25} \\ \hline
\multicolumn{11}{l}{\cellcolor[HTML]{EFEFEF}\textit{(3) Commercial API}} \\ \hline
Google STT&-&-&-&-&20.38&57.09&29.45&23.45&71.89&31.10 \\
\hline
\end{tabular}
\end{adjustbox}
\caption{\textbf{ASR evaluation results.} 
We report CER, WER, and sWER, where lower values indicate better performance. The table consists of three model groups: (1) zero-shot Whisper models of varying sizes, (2) Whisper-large-v2 fine-tuned on different datasets, and (3) a commercial ASR system (i.e., Google STT). In Group (2), \textit{General} refers to fine-tuning on KsponSpeech, and \textit{Weather$^*$} denotes fine-tuning on the TTS-generated weather dataset. The lowest and second-lowest scores of each column are highlighted in bold and underlined, respectively.}
\vspace{-0.2cm}
\label{tab:table_main}
\end{table*}

\subsection{Dataset Analysis}
\label{sec:data-analysis}

Table~\ref{tab:statistics} shows the overall statistics of our dataset, which includes 5,500 spoken queries. The average utterance length is 7.05 seconds, with an average of 7.59 words and 24.49 characters per query. Notably, 24.86\% of the unique words in our dataset are absent from the general-domain ASR corpus \citep{bang2020ksponspeech}, highlighting the specialized vocabulary of meteorological speech.

Table~\ref{tab:frequent} lists the most frequent words in general-domain and weather-domain datasets. While general speech contains common conversational terms, our dataset prominently features specialized meteorological vocabulary. This contrast underscores the need for domain-specific datasets to improve ASR performance in technical applications.

\section{Experimental Setup}
\subsection{Evaluation Datasets}
We use our Korean ASR dataset for the weather domain to evaluate different ASR models.
The dataset comprises 5,500 spoken queries, which we randomly split into 500 samples for the development set and 5,000 samples for the test set.
Additionally, we utilize the \textit{eval-clean} test set from the \textit{KsponSpeech} dataset \citep{bang2020ksponspeech} to evaluate the models in a general conversation domain.
This dataset consists of 3,000 audio files and their corresponding answer transcriptions.
% This approach allows us to compare models across both domain-specific and general ASR tasks.

\subsection{Metrics}
We use Character Error Rate (CER) and Word Error Rate (WER) as evaluation metrics to measure the distances between the prediction and ground-truth transcription.
Additionally, we employ a space-normalized Word Error Rate (sWER) \citep{bang2020ksponspeech}, which accounts for the flexibility and variations of space rules in Korean. 

\subsection{Models and Training Datasets}
\noindent{\textbf{Zero-shot Multilingual Models}} For our experiments, we used the multilingual Whisper model family \citep{whisper}, evaluating four different model sizes: \texttt{tiny} (39M), \texttt{small} (244M), \texttt{medium} (769M), and \texttt{large-v2} (1550M). These pre-trained models were assessed in a zero-shot manner to evaluate their performance without additional fine-tuning on our dataset. The target language and task in Whisper’s prefix tokens were set to \textit{Korean} and \textit{transcribe}, respectively. 
% Additionally, for evaluation in the weather domain, we tested Google's commercial speech-to-text API\footnote{\url{https://cloud.google.com/speech-to-text}}.

\noindent{\textbf{Fine-tuning on General Open-domain Dialogues}} To explore the impact of adapting a multilingual ASR model to the Korean language, we fine-tuned an ASR model using the KsponSpeech \citep{bang2020ksponspeech} dataset.
The dataset contains 619k training instances about open-domain dialogue utterances from Native Korean annotators. 
The Whisper large-v2 is fine-tuned on this dataset.
More implementation details are in Appendix \ref{ref:implementation_details}.

\noindent{\textbf{Data Augmentation for Meteorological Domain}}
Inspired by \citet{zheng2021using}, we hypothesize that teaching the model the pronunciation of specialized weather terms is crucial for accurate transcription. To this end, we used a TTS system to generate audio for 10k and 9.8k natural language queries from the URL and SQL subsets of \citet{jo-etal-2023-integrated}, respectively. Queries in our evaluation set were excluded to prevent test leakage. We employed the Google TTS service\footnote{\url{https://gtts.readthedocs.io/}} and converted English words to Korean pronunciations based on expert guidelines (Section~\ref{sec:data-pipeline}) before synthesis. The resulting TTS audio was used to fine-tune Whisper large-v2, either alone or combined with KsponSpeech, yielding two ASR variants.

\noindent{\textbf{Post-processing with Unimodal LLMs}}  
To assess whether large language models (LLMs) can improve ASR outputs without acoustic input \citep{chen2023hyporadise, hu-etal-2024-listen}, we used GPT-4o-mini \citep{achiam2023gpt} to revise the top-1 transcription from Whisper-large-v2. The model was evaluated in both zero-shot and 20-shot settings, using examples from the development set. We adapted the prompt from \citet{chen2023hyporadise} to the Korean meteorological domain to better handle domain-specific terms. This setup allows us to test how effectively an unimodal LLM can refine ASR transcriptions based solely on textual input.

\subsection{Implementation Details}
\label{ref:implementation_details}
All fine-tuned models are parameter-efficiently trained using LoRA \citep{hulora} with $r$=32.
The fine-tuning process involved 3 epochs of training with a batch size of 48, a learning rate of 1e-3, and a warmup ratio of 0.1 with AdamW optimizer \citep{loshchilovdecoupled}. 
For fine-tuned models, we saved the checkpoint after each epoch and selected the best one based on the lowest CER score from the development set of our dataset.
The greedy decoding \citep{holtzmancurious} is used as a decoding algorithm for all models.
All models are implemented with the Transformers framework \citep{wolf-etal-2020-transformers} and PyTorch \citep{paszke2019pytorch}.
For the post-processing with unimodal LLMs,  we used the following text prompt: \textit{"Below is the best-hypotheses transcribed from speech recognition system for Korean Meteorological experts. Please try to revise it and write the response for the true transcription."}

\section{Results and Analyses}
% Table~\ref{tab:table_main} summarizes results across different setups.
In this section, we describe the key observations. Table~\ref{tab:table_main} summarizes results across different setups.

\paragraph{Zero-shot ASR models struggle with domain-specific terminology}
Whisper models show high CER and WER when transcribing meteorological queries, particularly for specialized terms. Larger models (e.g., Whisper-large-v2) reduce errors relative to smaller variants but still struggle with domain-specific vocabulary, highlighting the need for further adaptation.

\paragraph{Fine-tuning improves general performance, but domain-specific adaptation is essential}
Fine-tuning Whisper-large-v2 on KsponSpeech lowers CER and WER overall, but improvements on meteorological queries are limited. This suggests that general-domain data alone is insufficient for handling specialized terms.

\noindent{\textbf{Synthetic data (TTS) aids domain adaptation}}
Training with TTS-generated weather-domain queries improves transcription of meteorological terms. The model fine-tuned on both KsponSpeech and synthetic data achieves balanced performance across general (CER 10.54\%) and weather domains (CER 17.46\%), while the model trained only on synthetic data achieves the lowest error rates in the weather domain (CER 10.64\%).

\begin{figure}[t!]
\centering
\includegraphics[width=\columnwidth]{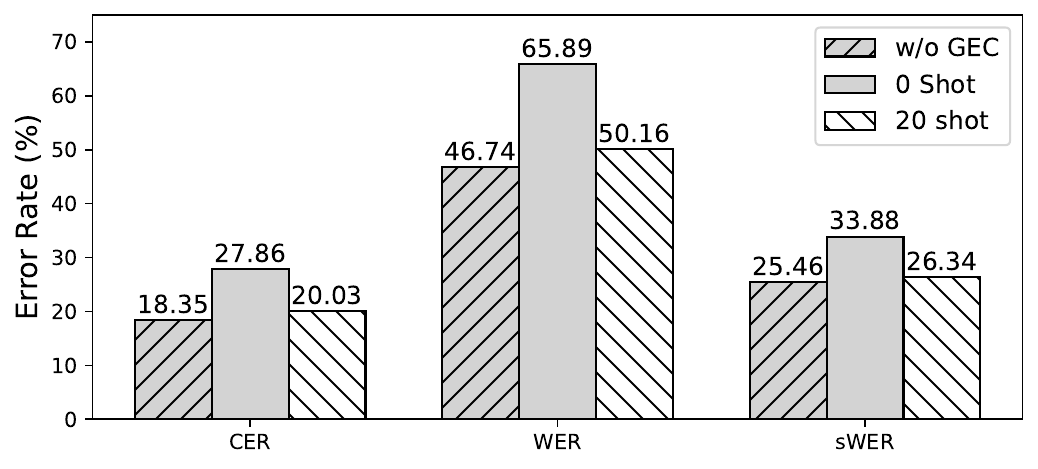}
\vspace{-0.8cm}
\caption{\textbf{Effect of LLM-based post-processing.}
We compare three conditions: without LLM-based grammatical error correction (\textit{w/o GEC}), with zero-shot (\textit{0 Shot}), and with 20-shot (\textit{20 Shot}) setups.
}
\vspace{-0.4cm}
\label{fig:gec}
\end{figure}

\paragraph{LLM-based post-processing yields inconsistent gains}
As shown in Figure~\ref{fig:gec}, LLM-based post-processing does not consistently improve recognition of domain-specific terms. Error rates remain higher than the zero-shot baseline, suggesting that text-only refinement is insufficient without acoustic adaptation. While training a dedicated GEC model or using N-best hypotheses \citep{chen2023hyporadise,hularge} may help, these approaches require large corpora and increase inference costs.

\paragraph{Commercial ASR systems perform poorly on specialized data}
Google’s STT model yields the highest error rates (e.g., 20.38\% on the Weather-test set), underperforming even the zero-shot Whisper-large-v2. This underscores the limitations of generic ASR systems in specialized domains and the need for targeted adaptation. Overall, these results highlight the importance of fine-tuning with domain-specific data. While LLMs can support transcription refinement, substantial improvements require acoustic-level adaptation.

\paragraph{Case Study}
To illustrate common ASR errors, we present examples from Table~\ref{tab:main_sample_tables}. One issue involves domain-specific term misrecognition. In a wind forecast transcription, "지상 시계열 바람장미 풍속계급별 판촉횟수 CALM 0 5m/s 순별 하순 211 인제" \textit{("Ground-level wind rose classified by wind speed, CALM 0–5m/s, late-stage 211 Inje")}, both zero-shot Whisper and Google STT misinterpret "CALM", distorting critical numerical information.
Another case involves unintended style shifts. In "8월 30일 증가압과 기온 평년값은 뭐니" \textit{("What are the historical averages for pressure and temperature on August 30?")}, the post-processed output rephrases it more formally as "8월 30일 증가압과 기온 평년값은 무엇인가요?". While fluency improves, such shifts may be undesirable in contexts requiring a colloquial tone. These cases reflect two key challenges: (1) preserving domain-specific and numerical terms, and (2) refining grammar while maintaining intended speech style. Addressing these requires domain-aware and context-sensitive correction strategies.

\begin{table}[t]
\centering
\small
\begin{adjustbox}{width=\columnwidth}
\begin{tabular}{ll}
\toprule
Answer & 8월 30일 증기압과 기온 평년값은 뭐니 \\
ZS & 8월 30일 증기압과 기\coloredsout{운 평량}값은 뭐니 \\
FT$_{General}$ & 8월 30일 \coloredsout{진}기압과 기온 평년값은 뭐니 \\
FT$_{Both}$ & 8월 30일 증기압과 기온 평년값은 뭐니 \\
Google STT & 8월 30일 증기압과 기온 평년 값은 뭐니 \\
LLM$_{20shot}$ & 8월 30일 증기압과 기온 평\coloredsout{량}값은 \coloredsout{무엇인가요} \\ \midrule
Answer & 관측횟수 CALM 0 5m/s 순별 하순 211 인제 \\
ZS & 관측\coloredsout{해수컴} 0 5m\coloredsout{f} 순별 하순 211 인\coloredsout{재} \\
FT$_{General}$ & 관측 \coloredsout{해수 칸 영} 5m \coloredsout{퍼셉커} 순별 하순 2\coloreduwave{11 인제}\\
FT$_{Both}$ & 관측\coloredsout{해서} CALM 0 5m\coloredsout{\%} 순별 하순 2 1 1 인제 \\
Google STT &  관측\coloredsout{에서 컴} 0 5 m\coloredsout{ 버서커} 순별 \coloredsout{화}순 2\coloredsout{일} 인제 \\
LLM$_{20shot}$ & 관측\coloredsout{해}수\coloredsout{컴} 0\coloredsout{$\sim$}5 m/s 순별 하순 211 인\coloredsout{재} \\

\bottomrule
\end{tabular}
\vspace{-0.1cm}
\end{adjustbox}
\caption{
\textbf{Qualitative Results} 
Incorrectly recognized characters are marked with \coloredsout{strikethrough}, while missing words are indicated with a \coloreduwave{wavy underline}. Space errors are omitted for better readability.}
\vspace{-0.5cm}
\label{tab:main_sample_tables}
\end{table}

\section{Conclusion}
We introduce a specialized evaluation dataset for assessing ASR performance in the Korean meteorological domain, addressing the lack of domain-specific benchmarks. Our experiments reveal that while fine-tuning on general-domain data improves overall accuracy, specialized terminology remains a major source of error. By releasing this dataset, we aim to support further research in developing robust, domain-adapted systems that better reflect the demands of real-world forecasting scenarios.

% This work constructs a specialized evaluation dataset for assessing ASR models in the Korean meteorological domain, addressing the lack of domain-specific benchmarks. Our results show that pre-trained ASR models struggle with meteorological terminology, and while fine-tuning on general-domain data improves accuracy, domain adaptation remains essential for handling specialized vocabulary. While our dataset provides a valuable benchmark, future work should expand it to cover more diverse speech conditions, such as spontaneous and noisy environments, to better reflect real-world forecasting scenarios. By making this dataset available, we aim to drive further advancements in domain-adapted ASR systems for meteorological applications.

\section*{Limitations}
While our study provides a targeted benchmark for ASR in the Korean meteorological domain, several limitations remain. First, the dataset consists of scripted queries recorded in clean environments, which may not fully represent the \textbf{acoustic variability found in real-world settings}. In practice, meteorologists often speak spontaneously, with disfluencies, hesitations, or overlapping speech, especially during live broadcasts or team discussions. Our dataset does not yet capture such spontaneous or noisy conditions.
Second, while we explored text-only post-processing using unimodal LLMs, we did not investigate more \textbf{advanced correction strategies with acoustic cues} or multiple ASR hypotheses. These approaches could further improve recognition of subtle or ambiguous terms but require additional data and computational resources.
Lastly, our work is focused on a single domain (meteorology) and a single language (Korean). The findings may not generalize to \textbf{other specialized domains}, such as healthcare or law, or to other low-resource languages. Expanding this research to cross-domain and cross-lingual settings remains an important direction for future work.

\section*{Ethical Statement}
All spoken queries in our dataset were recorded with the informed consent of native Korean speakers, who were compensated fairly for their participation. The recordings were collected in controlled environments to ensure quality and to minimize any unintended background content. To protect speaker privacy, no personally identifiable information (PII) was included in the dataset. All utterances were manually reviewed to avoid harmful, offensive, or culturally insensitive content. While the dataset aims to represent meteorological language use in Korean, we acknowledge that it may not fully capture regional or situational variations in speech patterns.

\section*{Acknowledgement}
This work is supported by a grant of a Developing Intelligent Assistant Technology and Its Application  for Weather Forecasting Process (KMA2021-00123), Institute for Information \& communications Technology Planning \& Evaluation(IITP) grant funded by the Korea government(MSIT) (RS-2019-II190075, Artificial Intelligence Graduate School Program(KAIST)), and the National Research Foundation of Korea(NRF) grant funded by the Korea government(MSIT) (No. RS-2025-00555621).

\bibliography{custom}

\appendix

\newpage

\section{Pronunciation Guideline Details for Speech Data Collection}
\label{app:sec:guideline}

\begin{figure}[t!]
\centering
\includegraphics[width=\columnwidth]{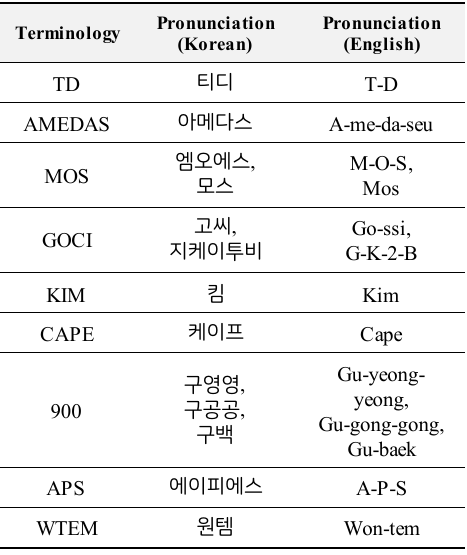}
\caption{Samples of pronunciation rules for meteorological terminologies. The \texttt{Pronunciation (English)} are manually written by the authors for clarity and were not provided at the data annotation phase.
}
\label{fig:guideline_example}
\end{figure}

To ensure accurate pronunciation of meteorological terms, we consulted domain experts from the National Institute of Meteorological Sciences (NIMS). 
The focus was primarily on English words and abbreviations, along with various unit and number expressions. 
We note that a single word or expression can be pronounced in different ways.
Experts provided detailed pronunciation guidelines for 256 words and expressions frequently used in the Korean weather domain.
These guidelines were given to annotators as an initial reference and were available throughout the recording process. 
Selected examples from the pronunciation guidelines are shown in Fig. \ref{fig:guideline_example}.

\section{Qualitative Results}
\begin{figure*}[t!]
\centering
\includegraphics[width=\textwidth]{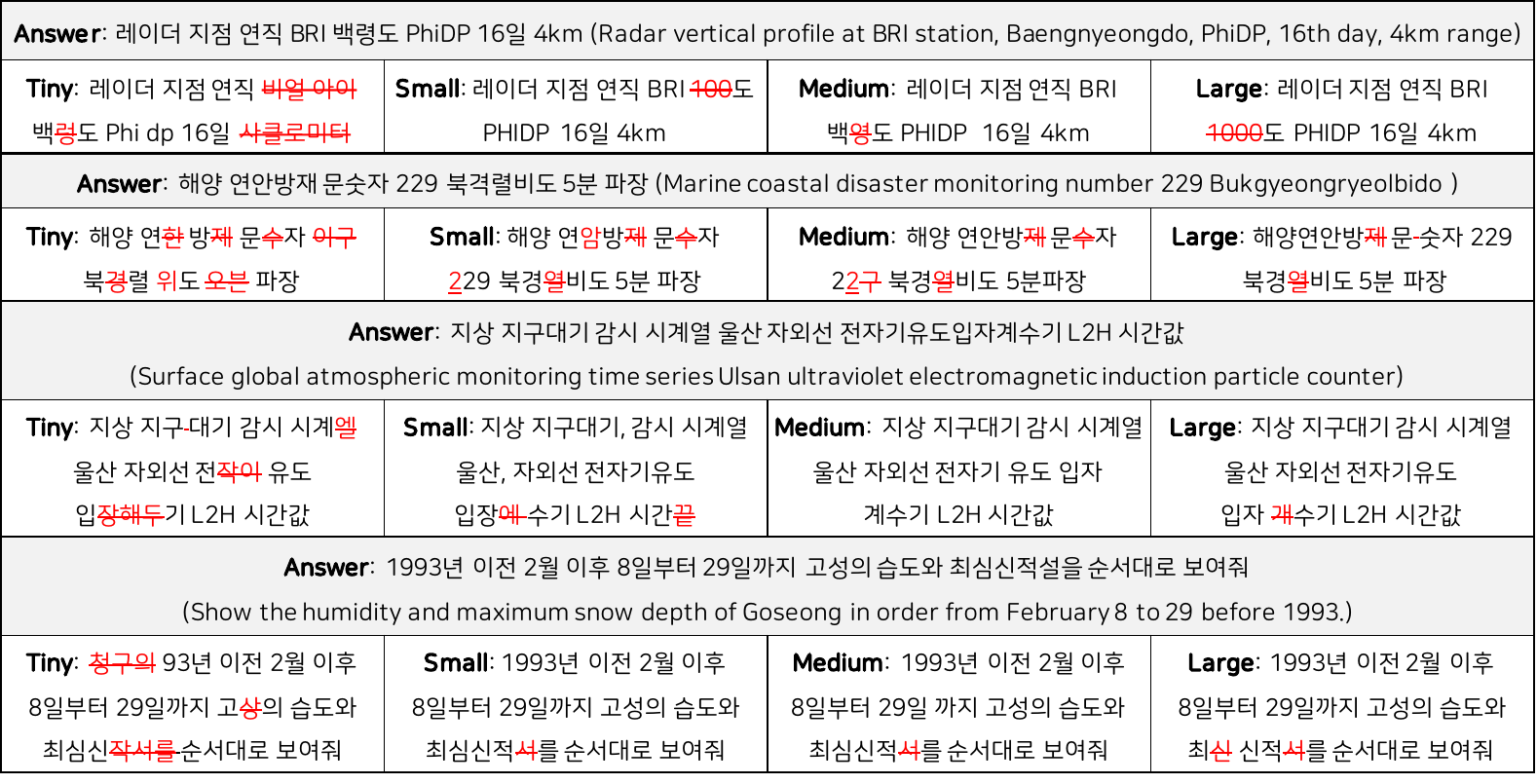}
\caption{\textbf{Qualitative results of pre-trained Whisper models with different model sizes.} 
The answer queries in English are manually translated from the original Korean answer.
Incorrect model predictions are highlighted with \coloredsout{strikethrough}. }
\label{fig:zs.case_study}
\vspace{-0.2cm}
\end{figure*}

\begin{figure*}[t!]
\centering
\includegraphics[width=\textwidth]{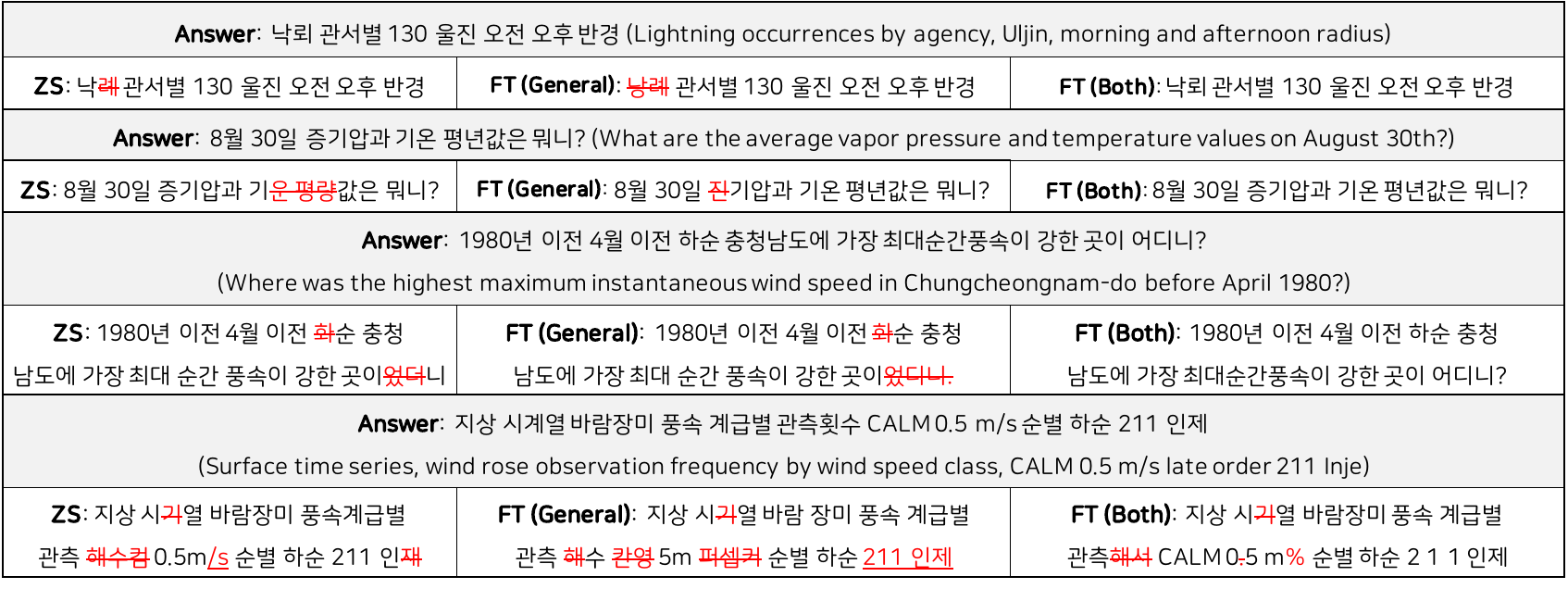}
\caption{\textbf{Qualitative results of Whisper models finetuned on different datasets.} 
All indicators are the same with Figure \ref{fig:zs.case_study}.}
\label{fig:ft.case_study}
\vspace{-0.2cm}
\end{figure*}

We present further prediction results of different open-source ASR models on Figures 
\ref{fig:zs.case_study} and \ref{fig:ft.case_study}.

\end{document}